\documentclass[conference]{IEEEtran}
\IEEEoverridecommandlockouts
\usepackage{cite}
\usepackage{amsmath,amssymb,amsfonts}

\usepackage{soul}
\newcommand{\drop}[1]{\textcolor{red}{#1}}
\renewcommand{\drop}[1]{}

\usepackage{multirow}
\usepackage{xcolor}
\usepackage{enumitem}
\usepackage{hyperref}

\usepackage{graphicx}
\usepackage{booktabs} 
\usepackage{pifont}
\usepackage{mathtools}
\def\smallerspacecaption{\vspace{-2mm}}

\usepackage[ruled, vlined, norelsize]{algorithm2e}
\SetKwInput{KwInput}{Input}
\SetKwInput{KwOutput}{Output}
\SetKwInput{KwInit}{Initialization}
\SetKwInput{Kwprocedure}{Procedure}
\usepackage{lipsum}
\usepackage{dblfloatfix}
\usepackage{algpseudocode}
\usepackage{bm}

\newcommand{\toolname}[1]{\texttt{ChipCoder}}

\definecolor{cadmiumgreen}{rgb}{0.0, 0.42, 0.24}

\usepackage{listings}
\lstdefinelanguage{Verilog}{
  morekeywords={module, endmodule, input, output, reg, wire, always, begin, end, if, else, for, while, case, default},
  sensitive=false,
  morecomment=[l]{//},
  morecomment=[s]{/*}{*/},
  morestring=[b]",
}
\definecolor{shadecolor}{rgb}{0.9,0.9,0.9}
\usepackage{acronym}

\begin{document}

\title{Free and Fair Hardware: A Pathway to Copyright Infringement-Free Verilog Generation using LLMs\thanks{*These authors contributed equally to this work.}}
\author{Sam Bush*, Matthew DeLorenzo*, Phat Tieu, Jeyavijayan Rajendran\\
Texas A\&M University, USA\\
{\tt \{samuelkbush, matthewdelorenzo, phat.tieu, jv.rajendran\}@tamu.edu}} 

\maketitle

\begin{abstract}
Limitations in Large Language Model (LLM) capabilities for hardware design tasks, such as generating functional Verilog codes, have motivated various fine-tuning optimizations utilizing curated hardware datasets from open-source repositories. However, these datasets remain limited in size and contain minimal checks on licensing for reuse, resulting in potential copyright violations by fine-tuned LLMs. Therefore, we propose an evaluation benchmark to estimate the risk of Verilog-trained LLMs to generate copyright-protected codes. To minimize this risk, we present an open-source Verilog dataset, \href{https://huggingface.co/datasets/SETH-TAMU/FreeSet-V1.0-LabUse}{\texttt{FreeSet}}, containing over 220k files, along with the automated dataset curation framework utilized to provide additional guarantees of fair-use Verilog data. We then execute an LLM fine-tuning framework consisting of continual pre-training, resulting in a fine-tuned Llama model for Verilog, \texttt{FreeV}. Our results indicate that \texttt{FreeV} demonstrates the smallest risk of copyright-infringement among prior works, with only a 3\% violation rate. Furthermore, experimental results demonstrate improvements in Verilog generation functionality over its baseline model, improving VerilogEval pass@10 rates by over 10\%.
\end{abstract}

\begin{IEEEkeywords}
Large Language Models, Copyright, Verilog
\end{IEEEkeywords}

\section{Introduction}
\label{sec:Introduction}

\begin{figure*}[t]
    \centering
    \includegraphics[width=\textwidth,trim={0 0cm 0 0cm},clip]{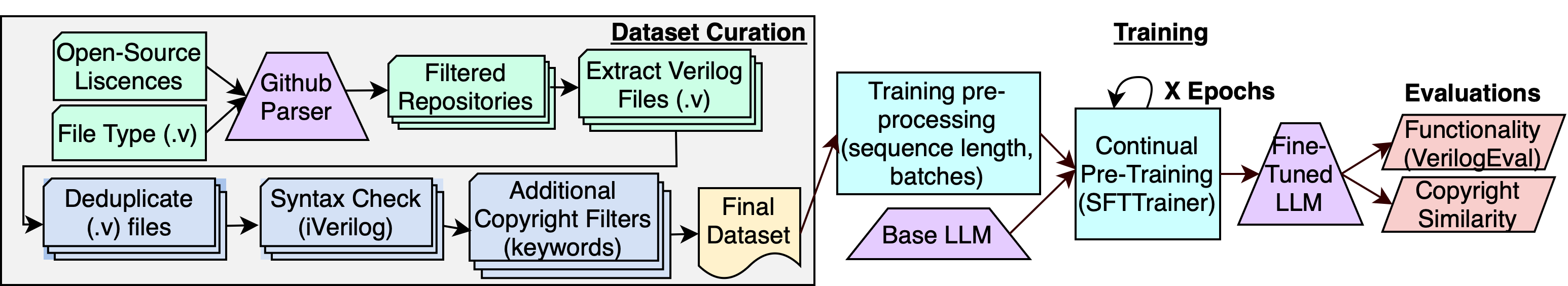}
    \caption{Framework - Dataset Curation and Continual Pre-Training}
    \label{fig:final_framework_figure}
\end{figure*}

Recent advancements in the domain of generative artificial intelligence (AI) have resulted in the development of Large Language Models (LLMs)~\cite{vaswani2023attention}. These models, consisting of billions of neurons composing deep neural network architectures~\cite{vaswani2023attention, wolf2020transformers}, have enabled the ability to effectively understand textual information and produce high-quality responses. These versatile capabilities, emerging from training procedures on vast amounts of text data, have led to their integration into a wide range of industries to automate workflows~\cite{LLMs_applications_blog}, specifically in code generation for software development~\cite{guo2024deepseek, chen2021evaluating, zhu2024deepseek}.
Significant efforts have sought further optimizations of LLMs for code generation, largely through increases in the scale and quality of their training datasets~\cite{kocetkov2022stack}, enhancing the syntax and functionality across a wide range of coding tasks~\cite{chen2021evaluating}. These efforts, spearheaded by Microsoft~\cite{stratton2024introduction}, Meta~\cite{roziere2023code}, Google, and OpenAI~\cite{achiam2023gpt}, have resulted in widely adopted code-generation models and tools.

Taking inspiration from LLM performance in software, an emerging application of LLMs is in the chip design process~\cite{zhong2023llm4eda}. This hardware design flow to develop integrated circuits (ICs) for use in electronic devices is becoming increasingly complex, error-prone, and expensive. Furthermore, an integral component of IC design and verification is enacted through code in hardware description languages (HDL), defining the behavior of digital circuit designs. This has motivated the utilization of LLMs in translating natural language hardware design requirements into register-transfer-level (RTL) code representations~\cite{thakur2024verigen, blocklove2023chip, chang2023chipgpt}. Research efforts have demonstrated optimizations in LLM for hardware design through a variety of approaches, including the inference process, fine-tuning the model itself on curated Verilog datasets~\cite{zhao2024codev, liu2024rtlcoder, thakur2024verigen, pei2024betterv}, reinforcement learning~\cite{wang2024large}, and frameworks integrating prompt and tool feedback~\cite{thakur2023autochip, blocklove2023chip, tsai2024rtlfixer}. As a result, the ability of LLMs to produce compilable and syntactically correct Verilog has gradually improved. 

Although these works demonstrate improvements to LLMs in the hardware domain, challenges remain within the fine-tuning process, limiting the opportunity for further LLM optimizations and practical adoption in hardware design. One is the availability of high-quality and real-world Verilog and RTL datasets for utilization in training, as compared to software languages, LLMs have a significantly smaller dataset corpus~\cite{thakur2024verigen, liu2024rtlcoder}. Furthermore, prior efforts to curate Verilog datasets on open-source repositories have minimal filters for identifying licensing or copyright restrictions of these repositories, risking the unauthorized reuse of proprietary code by fine-tuned models. This is a pertinent concern for commercial LLMs, as OpenAI currently faces a lawsuit in which CoPilot is claimed to violate copyright law through open-source training procedures~\cite{openailawsuit2023}. Furthermore, emerging threats of hardware design IP piracy~\cite{5401214} by LLMs can also be exacerbated through improper use of data in training and fine-tuning.

To overcome this challenge, we first (i) provide an open-source inference benchmark that evaluates the proclivity of Verilog-tuned LLMs to generate copyright-protected hardware designs based on a similarity threshold. (ii) We then propose an automated framework that enables the large-scale extraction of open-source Verilog files from Github, resulting in a curated high-quality Verilog dataset of over 220k diverse files, surmounting over 16GB of text data. Notably, unlike prior open-sourced datasets, this framework checks for permissive and non-permissive licenses on each repository along with file-by-file identification of copyright keywords to remove any codes with a high likelihood of copyright protection, minimizing the risk of LLM copyright infringement. (iii) This dataset is then utilized to perform continual pre-training on a state-of-the-art local LLM (\texttt{Llama-3.1-8B-Instruct}), enabling the functional performance and copyright similarity of the fine-tuned model to then be evaluated. 

Our results demonstrate that ~\texttt{FreeV} has the smallest potential for hardware copyright infringement among prior works, with only a 3\% violation rate. Furthermore,~\texttt{FreeV} demonstrates improvements over its base Llama model when evaluated on VerilogEval, increasing the pass@5 rate by 7.9\%. and the pass@10 rate by 10.1\%.

Therefore in this paper, we provide the following contributions:

\begin{enumerate}

\item A novel benchmarking procedure in which the risk of copyright infringement can be estimated in fine-tuned LLMs for hardware generation by similarity analysis.
\item To the best of our knowledge, we devised the first open-source Verilog dataset and curation framework for LLM fine-tuning that checks for both licenses and copyright content in each file, minimizing copyright risk. The dataset is available here: \href{https://huggingface.co/datasets/SETH-TAMU/FreeSet-V1.0-LabUse}{https://huggingface.co/datasets/SETH-TAMU/FreeSet-V1.0-LabUse}.
\item A fine-tuned Llama-3.1-8B model on our Verilog dataset \texttt{FreeV}, demonstrating relative improvements in Verilog generation over its baseline and minimal hardware copyright violations.
\end{enumerate}
\section{Background and Related Work}
\label{sec:Background}

\subsection{LLM Risks in Copyright Infringement}
\label{sec:LLM_risks}
As LLMs continue to scale in size and training datasets expand, the ability of LLMs to learn, memorize, and respond with texts that are copyright-protected is becoming an increasing risk to state-of-the-art models. Questions regarding the legal implications of the definition of ``fair use" in model training and generation are actively debated~\cite{meeus2024copyright, rahman2023beyond}. This has given rise to a series of lawsuits facing prominent generative-AI entities, including Stability AI in which plaintiffs allege improper use of their image content in training~\cite{samuelson2023ongoing}, along with claims by the New York Times that GPT-4 infringes copyright by reproducing articles in its output~\cite{freeman2024exploring}. This has also impacted LLMs for code generation, in which Microsoft and OpenAI are claimed to have improperly utilized and monetized open-source programs~\cite{openailawsuit2023}. These developments have motivated the need to develop frameworks to identify if copyrighted content in utilized in LLM training~\cite{li2024digger} and prevent the generation of copyrighted text~\cite{liu2024shield}.\\
This concern of improper use of code in training is relevant to the hardware domain regarding IP piracy as LLMs begin to be fine-tuned on hardware datasets. Beyond utilizing open-source code and preliminary license checks~\cite{pei2024betterv}, minimal protections have been established in ensuring copyright-protected hardware is not utilized in LLM training.

\subsection{LLM Fine-Tuning for Hardware Design}
To effectively utilize emerging fine-tuning methodologies available for LLM optimization towards Verilog code generation tasks, a sufficiently sized and high-quality curated dataset is essential. Many works explore prompting state-of-the-art models (including GPT) for various RTL generation tasks~\cite{lu2024rtllm, blocklove2023chip, chang2023chipgpt, kande2023llm}. However, privacy limitations associated with API interfaces in commercial LLMs and limited optimization for specific use cases~\cite{liu2024openllm} motivate the use of open-source LLMs. Initial works such as VeriGen~\cite{thakur2024verigen} first explored fine-tuning local models on a curated dataset of Verilog texts from GitHub and textbook data sources, demonstrating up to 41\% improvement in Verilog code-generation tasks after additional fine-tuning on the dataset itself.

Furthermore, RTLCoder~\cite{liu2024rtlcoder} proposed a methodology to leverage LLMs to supplement the Verilog dataset through LLM-assisted generation, in which Verilog codes are generated by GPT through a mutation-based framework. This work additionally generated English descriptions of the module, enabling instruction fine-tuning, in which LLMs are trained on ideal prompt and response structures for further LLM alignment. BetterV~\cite{pei2024betterv} proposes to augment the instruction-tuning Verilog dataset by pairing Verilog modules with C translations, enabling the model's existing knowledge of C to be leveraged. CodeV~\cite{zhao2024codev} further expands Verilog datasets for instruction tuning by scraping GitHub for syntactically correct Verilog modules and prompting an LLM to generate effective summarizations of the code. Furthermore,~\cite{wang2024large} proposes the utilization of reinforcement learning through PPO, in which rewards are provided as feedback based on a golden hardware solution. To the best of our knowledge, only BetterV performs repository license filters in their curation process, with none indicating copyright checks on a file-by-file basis.

In addition to fine-tuning, researchers have proposed alternative methods of optimizing LLM performance in hardware design. This includes ChipChat~\cite{blocklove2023chip}, which demonstrates that through extended conversational prompting frameworks, larger-scale hardware designs can be effectively generated. Feedback from tools such as human and EDA feedback are additionally proposed in~\cite{blocklove2023chip, chang2023chipgpt}. Optimizations to the inference (token-selection) process itself can be seen in~\cite{delorenzo2024make}, where Monte-Carlo Tree Search enables iterative feedback for PPA and functionality optimization. Additional methodologies are proposed in~\cite{pei2024betterv} through discriminator guidance.
\section{Framework}
\label{sec:Framework}
\begin{table*}[htbp]
    \centering
    \small
    \caption{Comparison of FreeSet with Prior Curated Hardware Design Datasets}
    \addtolength{\tabcolsep}{-1pt}
        \begin{tabular}{lllllll} 
            \toprule
            LLM & Size (Disk) & Size (Rows) & Dataset Structure & Augmented (Y/N) & Open-Source & License Check\\
            \midrule
            VeriGen's Dataset~\cite{thakur2024verigen} & 1.89 GB & 108,971 & Continual Pre-Training & No & Yes & No\\ 
            RTLCoder~\cite{liu2024rtlcoder} & 55.1 MB & 27,000 & Instruction-Tuning & Yes & Yes & No \\
            CodeV~\cite{zhao2024codev} & N/A & 165,000 & Instruction-Tuning & Yes & No & No\\
            BetterV~\cite{pei2024betterv} & N/A & N/A & Instruction-Tuning & Yes & No & Yes\\
            CraftRTL~\cite{liu2024craftrtl} & N/A & 80,100 & Instruction-Tuning & Yes & No & No\\
            OriGen~\cite{cui2024origen} & 548 MB & 222,075 & Instruction-Tuning & Yes & Yes & No\\
            \textbf{FreeSet (This work)}  & \textbf{16.5 GB} & \textbf{222,624} & \textbf{Continual Pre-Training} & \textbf{No}& \textbf{Yes} & \textbf{Yes}\\
            \bottomrule
        \end{tabular}
    \label{tab:dataset_compare_table}
\end{table*}

\begin{figure}[t]
    \centering
    \includegraphics[width=0.48\textwidth]{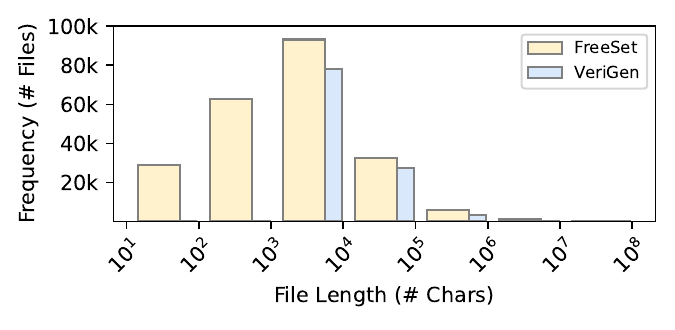}
    \smallerspacecaption
    \smallerspacecaption
    \smallerspacecaption
    \caption{Distribution in Verilog File Lengths of FreeSet vs. VeriGen~\cite{thakur2024verigen}}\label{fig:impact_of_baseline_reward} 
\end{figure}

\subsection{LLM Hardware Copyright Infringement Benchmark}
\label{sec:frameworkA}
We first examine the validity of the assumption that prior works training LLMs on Verilog datasets can risk copyright infringement through reproducing licensed code. Several LLMs have been trained using open-source code on GitHub and other code-sharing platforms~\cite{thakur2024verigen,pei2024betterv,zhao2024codev}. Although they are trained on open-source material with some license checks, some files can contain copyrighted and proprietary materials.

To demonstrate this hypothesis, we employed prompt engineering with the intention of extracting copyright training data from those LLMs. If these models generate copyrighted Verilog codes to a certain similarity threshold, the original training data for these models is likely to contain such copyrighted data. First, we curated a dataset with copyrighted materials that is available on GitHub. This dataset, collected in Section~\ref{sec:frameworkB}, identifies copyright declarations in files often through comments and isolates the Verilog modules themselves for use in this benchmark. The result was a collection of 2K files with copyright licenses from several companies, such as Intel and Xilinx. Next, we used the text in these copyrighted materials as the input prompts for the LLMs. As these files still contained copyright-related information in the comments, all comments inside the files were stripped, leaving only the code behind. We then curated a set of 100 input prompts. Each prompt contained the first 20\% of a copyrighted code file, with a limit of 64 words per prompt.

After feeding each prompt to the LLMs, the output was compared against the copyrighted dataset using cosine similarity. This score ranges from 0 to 1, where 0 represents no similarity, and 1 signifies an exact match. We use a threshold of 0.8 or greater to determine whether the output originated from the copyrighted dataset.

\subsection{Dataset Curation}\label{sec:frameworkB}
With the benchmark established to identify the risk of copyright infringement, we now seek to create a comprehensive dataset of varying file sizes that additionally ensures copyright-protected codes are not included. The framework (Figure. \ref{fig:final_framework_figure}) for generating this dataset is described below.
\subsubsection{\textbf{Challenge 1. Extracting Verilog Data}} One limitation is the ability to effectively extract significant amounts of Verilog files from online sources. Prior works have utilized tools including Google BigQuery, GitHub sources, and hardware design data extracted from textbooks, along with a series of augmentation techniques to expand a base set of data through translation~\cite{pei2024betterv} or mutations~\cite{liu2024rtlcoder}. Furthermore, this work solely focuses on extracting and isolating high-quality data from real-world, open-sourced, and open-licensed sources, resulting in the utilization of data scraping tools. Furthermore, these APIs consist of their own challenges, including costs, throughput bottlenecks from rate limitations, and granularity.\\
\subsubsection{\textbf{Solution 1}} This challenge is addressed through the use of an automated system (implementing the GitHub API) to efficiently scrape repositories and extract Verilog code. This enables the dataset to including up-to-date Verilog code, as resources such as the Google BigQuery GitHub dataset used for VeriGen~\cite{thakur2024verigen} have not been updated since 2022. At the time of curation, there are \(\sim\)50K publicly available GitHub repositories containing Verilog data, with a total of 1.3 million Verilog files. The GitHub API is limited to only being able to display 1K results per query for non-enterprise accounts. To avoid hitting this limitation, the system was designed to granularize queries by repository creation date ranges, from 2008 when GitHub was established to 2024. Further granularization was performed by filtering for specific repository licenses, reducing the results per query significantly.

The resulting repositories in each query are then cloned to gather all of their data and author information for proper accreditation. Most files in the repositories were miscellaneous and consisted of non-Verilog data (i.e. readme files, binary test data, compiling information), so the Verilog data within the cloned repositories would then be filtered for and condensed into a large bank of Verilog files.

\subsection{Removing Copyright-Protected Code}
\label{sec:frameworkC}
\subsubsection{\textbf{Challenge 2. Removing Copyright Protected Codes}} An additional challenge using data gathered from GitHub is a lack of moderation regarding copyright restrictions. While GitHub enables massive sources of open-source data, many repositories lack proper licensing to ensure their data is not proprietary, and code bases under open-source licensing may still contain proprietary or copyrighted information. While most Verilog datasets from similar works~\cite{zhao2024codev, liu2024rtlcoder} that gather data from GitHub repositories report it to be open-source, the details and steps gone through to verify the data gathered are either non-existent or not comprehensive. This framework therefore implements additional precautions to ensure there is no data under private copyright within the training dataset.

\subsubsection{\textbf{Solution 2}} As part of the process of granulating queries, a select set of commonly used open-source licenses, both permissive and non-permissive: MIT, Apache 2.0, variations of the GNU General Public License and GNU Lesser GPL, Mozilla Public License, Creative Commons License, Eclipse Public Licenses, and BSD Licenses. While the intention of GitHub is for all repositories to be open-source, there are many repositories without a license on the repository that fall into a gray area in which they could potentially be part of a copyrighted code-base. As a result, these repositories were not considered for this dataset.

To ensure the dataset does not contain explicitly copyrighted data, a file-by-file filtering process is run to check the header comments of individual files for license information and combinations of language indicating the code to be under private copyright with combinations of keywords such as: ``proprietary," ``confidential," and ``all rights reserved," ensuring that only epositories with the fair-use modules are utilized. In the de-duplicated version of the dataset (described in Section~\ref{sec:filtration}), this filter was able to locate over 2K files under proprietary copyright licenses from companies such as Intel, some even containing possible encryption keys and other critical information. As this data was contained within reportedly open-source code bases, all datasets that gathered data under the impression that none of the data on GitHub would be under private copyright would be at risk for including highly proprietary information.

\subsection{Dataset Filtration}

\subsubsection{\textbf{Challenge 3. Ensuring quality of dataset}}\label{sec:filtration}When handling massive amounts of files, there are many potential issues and some considerations necessary to ensure quality. Often data, especially that of open-source files or recognized standards, will be reused heavily across thousands of files. This work will follow similar steps outlined in similar works (VeriGen~\cite{thakur2024verigen}) for handling duplicated or reused data within the dataset. Another potential challenge arises from syntax or compiler errors within the training set code which can effectively train errors into a model trained on the set. Further, prior works have also found challenges with biased data in a training set containing data found in testing frameworks, this work follows similar methods to solve these issues.

\subsubsection{\textbf{Solution 3}} The challenge of duplicated data within the set will bring the issue of over-fitting on potential models fine-tuned on the dataset, therefore de-duplication is required to clean the dataset for effective use. Following the method of de-duplication described in VeriGen, this work will implement a combined form of MinHash and Jaccard similarity. A method of Locality Sensitive Hashing (LSH) was implemented using MinHash signatures to represent each file. A similarity threshold of 0.85 was used to discard duplicate files, and LSH allowed for efficient querying for similar files in the dataset.

To additionally curate the dataset, the correctness of the syntax is then verified for each Verilog file as done in prior works~\cite{zhao2024codev, pei2024betterv}. We implemented this through the use of ~\textit{Icarus Verilog 10.3} to test the compilability of each code. As many files contain additional dependencies to modules outside of the files, only syntax-specific errors were identified and removed from the dataset in training.

\begin{table*}[ht!]
\caption{Comparison of LLMs Capabilities on VerilogEval}
\label{tab:functionality_results}
\centering
\begin{tabular}{ |c|c|c|c||c|c|c|  }
 \hline
 \multirow{3}{*}{\centering Type} & \multirow{3}{*}{Model} 
 & \multirow{3}{*}{Open-Source} & \multirow{3}{*}{Size} 
 &  \multicolumn{3}{|c|}{VerilogEval}\\
 \cline{5-7}
 & & & & \multicolumn{3}{|c|}{Human (\%)}\\
 \cline{5-7}
 & & & & Pass@1 & Pass@5 & Pass@10\\
 \hline
\multirow{4}{*}{\centering Foundation Models} & 
GPT-4 & No & N/A & 43.5 & 55.8 & 58.9\\
 & Codellama (CL) & Yes & 7B & 18.2 & 22.7 & 24.3 \\
 & DeepSeek-Coder (DS) & Yes & 6.7B & 30.2 & 33.9 & 34.9\\
  & CodeQwen (CQ) & Yes & 7B & 22.5 & 26.1 & 28.0\\
 \hline\hline
 \multirow{6}{*}{\centering Verilog-Tuned Models} & VeriGen & Yes & 16B & 30.3 & 43.9 & 49.6\\
 & RTLCoder-DS & Yes & 7B & 41.6 & 50.1 & 53.4\\
 & BetterV-CodeQwen & No & 7B & 46.1 & 53.7 & 58.2\\
  & CodeV-CodeQwen & Yes & 7B & 53.2 & 65.1 & 68.5\\
 & OriGen-DS & Yes & 7B & 54.4 & 60.1 & 64.2\\
   & CraftRTL-StarCoder2 & No & 15B & 68.0 & 72.4 & 74.6\\
  & OpenLLM-RTL & N/A & 6.7B & 42.8 & 51.6 & 55.0\\
 \hline\hline
  \multirow{2}{*}{\centering This Work} &
  Llama-3.1-Instruct (4-bit) & Yes & 8B & 14.8 & 23.0 & 25.9 \\
  & FreeV-Llama3.1 (4-bit)& Yes & 8B & 15.5 & 30.9 & 36.0
  \\
 \hline

\end{tabular}
\end{table*}

\subsection{Continual Pre-Training}
\label{FrameworkE}
\subsubsection{\textbf{Continual Pre-Training Setup}}
To perform continual pre-training using our dataset above, the SFTTrainer class from the Huggingface TRL library was utilized to implement the procedure in Python. We utilized the \texttt{Llama-3.1-8B-Instruct} as our base model for fine-tuning. The models were trained over 1 epoch across the full dataset, utilizing a single NVIDIA A100-40GB GPU. The training hyperparameters are defined as follows: max sequence length of 2048 tokens, and a training batch size per device of 16 with 2 gradient accumulation steps. Furthermore, to accommodate GPU memory limitations, QLoRA was applied to the model. This resulted in the quantization of the model's weights to 4-bit values and a LoRA layer with rank and alpha values equal to 8. Additionally, Unsloth was applied to accelerate the training process and minimize memory utilization.

\subsubsection{\textbf{Verilog Functionality Evaluation}}\label{sec:functionality_process}
To maintain consistency with the evaluations of prior works, we utilize the \textit{VerilogEval-Human 1.0.0} benchmark to measure the Verilog functionality rates of our fine-tuned LLM. This benchmark consists of 156 Verilog generation problems with human-generated descriptions of the intended Verilog module. As done in prior works, the success of the LLM is defined by the pass@k metric below, specifying the proportion of Verilog modules that can be generated functionally correct at least once in \textit{k} attempts from within a set of \textit{n} total generations:
\begin{equation}
\text{pass@k} = \underset{\text{problems}}{\mathbb{E}}\left[\frac{1-(\genfrac{}{}{0pt}{}{n-c}{k})}{(\genfrac{}{}{0pt}{}{n}{k})}\right]
\end{equation}
The inferencing process was then performed on the base model \texttt{Llama-3.1-8B-Instruct} and our fine-tuned model, FreeV in their 4-bit quantized representations. This is due to memory limitations in our inferencing process, in which we utilize an NVIDIA A5000 24GB GPU. For both models, the temperature, or randomness of the token selection, was set to 0.2 and 0.8, and the best result was chosen to follow the precedence set by prior works. Furthermore, the inferencing process limits token generation to 2048 and is set to end at the first instance of ``endmodule". The prompt for each VerilogEval definition simply consists of the VerilogEval-Human English description of the prompt, followed by the VerilogEval-Human prompt, consisting of the module instantiation, on the next line.

\begin{figure}[t]
    \centering
    \includegraphics[width=0.48\textwidth]{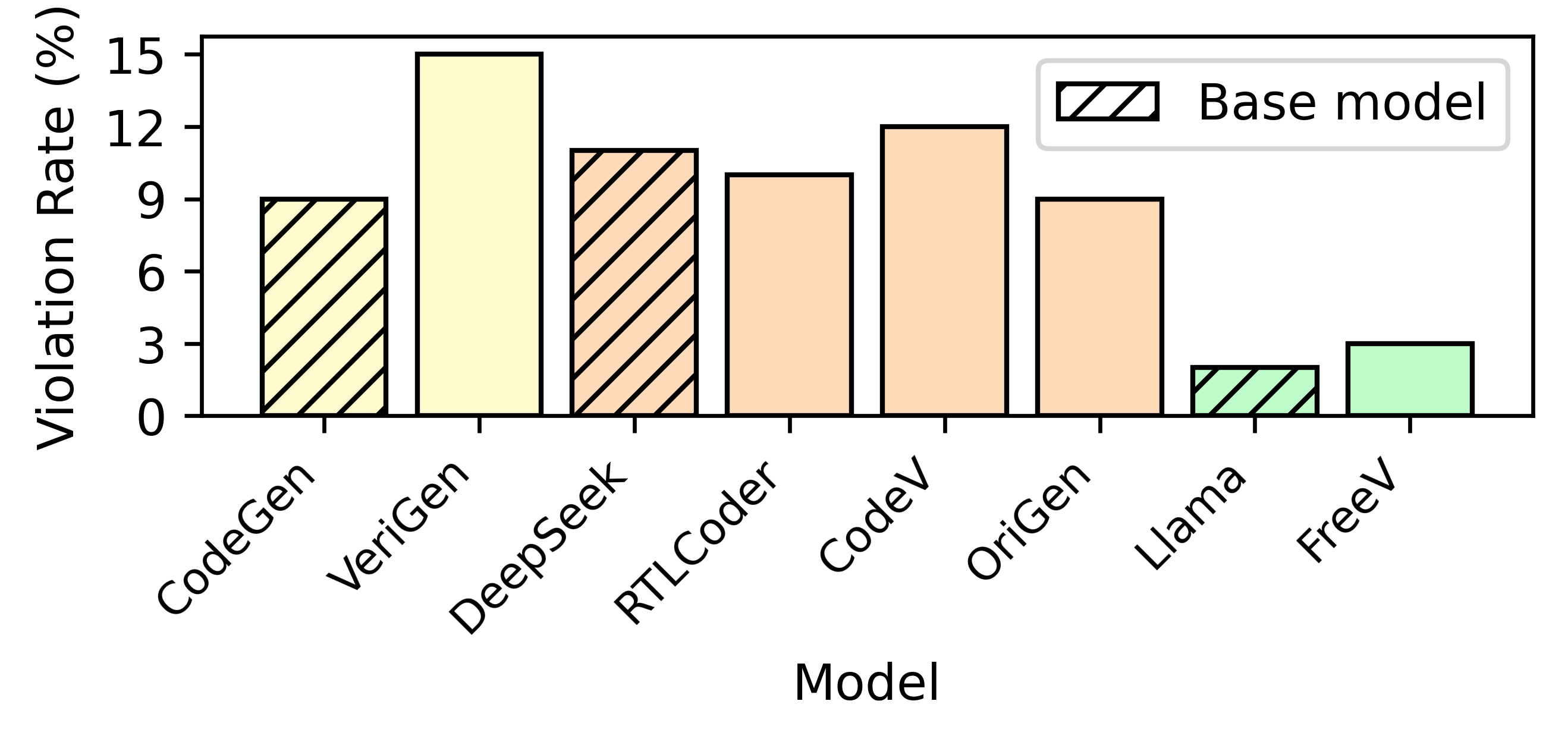}
    \caption{Hardware Copyright Infringement Rates across LLMs}
    \label{fig:copyright}
\end{figure}

\section{Experimental Evaluation}
\label{sec:results}
\subsection{Evaluation of Dataset}

The dataset configuration process described above gradually minimizes the dataset from its initial configuration. Through this minimization process, we can understand how each step impacts the dataset size. In the initial extraction of Verilog files from the repositories, over 1.3 million files were obtained. However, after extracting only items that have our set of open licenses, only 608,180 files remain. With this set of data, we can then perform LSH de-duplication which removes 62.5\% of the files. Finally, syntax and copyright checks were performed, resulting in a final dataset of 222K files. Through this process, we can find that the copyrighted data made up nearly 1\% of the original dataset. Additionally, we can determine that certain steps such as de-duplication and copyright checking, are crucial to minimizing the dataset size.

With the dataset curated above, we can then compare its distribution of files in terms of character count to gain insight into the diversity of the dataset. This is shown in Figure~\ref{fig:impact_of_baseline_reward}, in which we provide a direct comparison against the open-source VeriGen dataset. Through this chart, we can see that our dataset contains a much higher frequency of smaller files, with the vast majority of files ranging from 10 to 10,000 characters. Furthermore, our dataset additionally contained some extreme outliers, including a file of over 90M characters. This diversity of data supports the dataset's adaptability to a range of fine-tuning applications, from extended context length pre-training with models of large context windows (1M+ tokens) to smaller instruction-tuning pairs.

Additionally, we demonstrate additional dataset metrics in comparison to other open-source repositories, shown in Table~\ref{tab:dataset_compare_table}. Through these results, we see that our dataset is currently the largest open-source dataset in terms of size and number of Verilog files. This can be in part due to the curation procedures by other works such as CodeV, which removes files that exceed 2096 characters. Additionally, our dataset is notably the only open-source repository that contains explicit checks for licenses and copyright protection in each file (Table~\ref{tab:dataset_compare_table}).

\subsection{Copyright Findings}
In this section, we evaluate the results of the copyright detection framework defined in Section~\ref{sec:frameworkA}. This evaluation (seen in Figure~\ref{fig:copyright}), was performed on our set of different LLMs to demonstrate the impact that fine-tuning LLMs on various hardware datasets has on the rate of hardware copyright violations (i.e., the proportion of tested copyright codes that exceed the similarity threshold). For each fine-tuned LLM, we also evaluate its associated base model, enabling the role of the fine-tuning process itself to be isolated in the induction of the violations, as information learned in pre-training of the underlying base model is a significant consideration in the generated results. The models used for the evaluation include \texttt{fine-tuned-codegen-6B-Verilog} with the base model \texttt{codegen-6B-multi},  \texttt{RTLCoder-Deepseek-v1.1}, \texttt{OriGen} and \texttt{CodeV-DS-6.7B} with their base model \texttt{deepseek-coder-6.7b-base}, and our fine-tuned model with its base \texttt{Llama-3.1-8B-Instruct}. 

Through these evaluations, we find that the fine-tuned models VeriGen and CodeV have a higher rate of copyright violations in comparison to their base models, with VeriGen increasing from 9\% to 15\%. This indicates that the fine-tuning process itself on the respective curated datasets plays a significant role in increasing the rates of generating copyright-protected hardware designs, with alternative methodologies and datasets potentially utilizing some of these copyright-protected codes in their fine-tuning dataset.

Furthermore, we can see that our fine-tuned model trained on hardware code without copyright protections has the smallest hardware infringement rate (3\%) out of all of the fine-tuned LLMs. Additionally, our model only depicts a 1\% increase over its base model Llama-3.1. This supports the notion that removing copyright-protected texts can significantly minimize the likelihood of generating these files in the inference process.

\subsection{Evaluation of LLM Performance}
After the continual pre-training is complete, the base \texttt{Llama-3.1-8B-Instruct} and our fine-tuned model \texttt{FreeV} are both evaluated on VerilogEval (Human) prompts through the methodology described in Section~\ref{sec:functionality_process}, and compared to the reported values from prior works~\cite{zhao2024codev, liu2024craftrtl}. These include a set of foundational (pre-trained) models, along with state-of-the-art Verilog-Tuned models (Table~\ref{tab:functionality_results}).

We can see that through continual pre-training on our curated dataset, ~\texttt{FreeV} demonstrates improvements in the ability to generate functional Verilog modules when compared to its base version. This is seen in improving the pass@1, pass@5, and pass@10 rates by 0.7\%, 7.9\%, and 10.1\%, respectively. However, we can also see that the current training process does not provide state-of-the-art performance in comparison to prior works, as~\texttt{FreeV} does not outperform other fine-tuning models. This is largely expected, as most other works implement additional fine-tuning beyond causal pre-training, including instruction-tuning for further alignment. Also, there may be limitations due to 4-bit quantization.

However, we highlight that relative improvements are seen through pre-training on~\texttt{FreeSet} when compared to the base \texttt{Llama-3.1-8B-instruct} model.
\section{Discussion and Future Work}
In this work, we seek to address the limitations of prior Verilog datasets with respect to copyright infringement of hardware designs. This is done through first establishing a preliminary evaluation benchmark that estimates how often fine-tuned LLMs generate a predefined set of copyright-protected hardware designs when prompted, with violations defined through a cosine-similarity threshold. Although this approach is feasible and easily implemented, future work can explore additional prompting methods to target certain texts that the LLM may have learned through fine-tuning. Furthermore, the dataset of copyrighted texts can be further expanded tests through additional curation efforts to increase the robustness of the final determination. Lastly, other similarity metrics may be explored for effective comparisons of the hardware design, such as evaluating the design structure, like GNN4IP~\cite{yasaei2021gnn4ip}. 

The work then proposes an automated dataset curation framework, resulting in the largest open-source dataset of Verilog files, while additionally providing filters for copyright-protected codes. Future works may look to expand the current real-world texts through data-augmentation techniques, or through providing natural-language descriptions of the files for instruction-tuning purposes. 

Lastly, we demonstrate an example fine-tuning procedure through continual pre-training on a base Llama model. Although this procedure successfully brought relative improvements over its baseline, future works may implement more extensive training parameters (epochs, batch size, larger LoRA adapters) to extract additional functionality. Furthermore, methodologies may adapt this dataset to their alternative training strategies (such as instruction-tuning or reinforcement learning), enabling their limitations to be tested.

\label{sec:Discussion}

\section{Conclusion}
\label{sec:Conclusion}
Large language models have shown promise in their application to the hardware domain field, specifically in the generation of RTL code. Recent works have demonstrated success in applying emerging LLM fine-tuning techniques to LLMs for the task of RTL generation, enabling state-of-the-art performance on RTL tasks. However, these works remain limited by the dataset size of the hardware designs used in training, along with minimal checks regarding copyright infringement requirements for the curated files, resulting in the risk of fine-tuned LLMs violating copyright protections.

To address this limitation, we first propose a preliminary benchmark formulation to estimate the risk that existing LLM trained on Verilog codes will generate copyright-protected texts to a certain similarity threshold, indicating the use of protected hardware designs in the dataset. In an effort to then minimize this risk, we propose an end-to-end automated framework for extracting large quantities of Verilog files from online sources like GitHub, while notably also filtering out any licensed or copyright-protected files. This framework results in a curated and open-source dataset~\texttt{FreeV} for Verilog codes resulting in 16.5 GB and over 222k files. This dataset was then utilized to fine-tune an 8B parameter Llama model through continual pre-training for hardware code generation. 

Our evaluations indicate that our model~\texttt{FreeV} demonstrates the lowest frequency of copyright-infringement (only 3\% of codes), while also providing Verilog functionality improvements over its base model, increasing VerilogEval pass@10 rates by over 10\%.

\section*{Acknowledgment}
The authors acknowledge the support from the Center for Secure Microelectronics Ecosystem – CSME\#210205.

\bibliographystyle{IEEEtran}
\bibliography{main.bib}

\vspace{12pt}

\end{document}